\documentclass[runningheads]{llncs}
\usepackage{graphicx}
\usepackage{multirow}
\usepackage{longtable}
\usepackage[english]{babel}
\usepackage[utf8]{inputenc}
\begin{document}
\title{A Comparative Study of Feature Types for Age-Based Text Classification\thanks{Supported by the grant of the President of the Russian Federation no. MK-637.2020.9.}}
%
%
\author{Anna Glazkova\inst{1}\orcidID{0000-0001-8409-6457} and
Yury Egorov\inst{1}\orcidID{0000-0001-7670-5283} \and
Maksim Glazkov\inst{2}\orcidID{0000-0002-4290-2059}}
\authorrunning{A. Glazkova et al.}
%
\institute{University of Tyumen, ul. Volodarskogo 6, 625003 Tyumen, Russia \email{a.v.glazkova@utmn.ru}, \email{yurij.a.egorov@gmail.com}\and
"Organization of cognitive associative systems" LLC, ul. Gertsena 64, 625000 Tyumen, Russia
\email{my.eye.off@gmail.com}}
\maketitle              
\begin{abstract}
The ability to automatically determine the age audience of a novel provides many opportunities for the development of information retrieval tools. Firstly, developers of book recommendation systems and electronic libraries may be interested in filtering texts by the age of the most likely readers. Further, parents may want to select literature for children. Finally, it will be useful for writers and publishers to determine which features influence whether the texts are suitable for children. In this article, we compare the empirical effectiveness of various types of linguistic features for the task of age-based classification of fiction texts. For this purpose, we collected a text corpus of book previews labeled with one of two categories -- children’s or adult. We evaluated the following types of features: readability indices, sentiment, lexical, grammatical and general features, and publishing attributes. The results obtained show that the features describing the text at the document level can significantly increase the quality of machine learning models.

\end{abstract}
\section{Introduction}
Nowadays, there are quite a lot of approaches to text classification according to document subjects, genre, author or according to other attributes. However, modern challenges in the field of natural language processing (NLP) and information retrieval (IR) increasingly require classification based on more complex characteristics. For example, it may be necessary to determine whether the text contains elements of propaganda or whether it has similar plot characteristics to other texts. One of such urgent and complex classification tasks is the division of literary texts into suitable for children and for adults. Age-based classification tools could find wide practical application. For instance, they would be useful in the personal selection of fiction or in filtering content not intended for children.

Despite the fact that many scholars considered the issue of text difficulty estimation, formally, text difficulty does not indicate the age of the intended reader. The question of whether the features describing text difficulty are suitable for age-based classification needs to be investigated. In addition, the severity of the features depends on the text genre. It is necessary to find out how these or those features are presented in the literary text and whether they contain information about the age audience of the text.

In this paper, we systematically evaluated different feature types on age-based classification task. In addition to popular text difficulty features, we consider special publishing attributes intrinsic fiction books, such as age rating score and abstract features. We collected the corpus of Russian fiction texts and applied two commonly used machine learning models, these are random forest (RF) and linear support vector classifier (LSVC). For comparison, we evaluated a transformer model based on RuBERT and a convolutional neural network (CNN) trained on Word2Vec embeddings. Finally, we evaluated feedforward neural network (FNN) trained on RuBERT text embeddings and age rating scores.

The LSVC model using a combination of a baseline and publishing attributes showed the best result of 95.77\% (F1-score). RuBERT achives 90.16\%. The FNN model combining RuBERT embeddings and age ratings showed 94.78\%. The results show that the features describing the text at the document level gives an advantage in case of long texts. Moreover, publishing attributes provide valuable information for the age-based classifier. We also found that some features used to determine text difficulty positively affect the quality of age-based classification.

The paper is organised as follows. In Section 2 we present a brief review of related works. Section 3 describes feature types evaluated in the paper. Section 4 contains the description of our dataset. Section 5 presents the structure of the models and the evaluation results. Finally, Section 6 is a conclusion.

\section{Related Works}

In the modern world, the constant growth of information resources gives rise to the need for filtering and ranking texts. One of the significant characteristics of a text is its complexity. The question of determining text difficulty naturally looks related to the task of age-based classification.

The task of estimating texts by complexity is not new. It appeared at the beginning of the last century in the context of evaluating the readability of educational texts. Further, during the XX century, researchers have proposed a number of tests to determine readability based on the quantitative characteristics of texts. Readability tests usually use quantitative text features, such as counting syllables, words, and sentences. There are several common readability texts for text difficulty estimation. For instance, these are: the Flesch–Kincaid readability test, the Coleman–Liau index, the automated readability index (ARI), the SMOG grade, the Dale-Chall formula \cite{Crossley,Didegan}. 

\begin{itemize}
    \item The Flesch–Kincaid readability test is based on the idea that the shorter the sentences and words, the simpler the text. The specific mathematical formula is: 
    \begin{equation}
      \textit{R}_\textit{F} = 206.835 - 1.015  \cdot \textit{ASL} - 84.6  \cdot \textit{ASW},
    \end{equation}
    \textit{ASL} -- average sentence length,
    \textit{ASW} -- average number of syllables per word (i.e., the number of syllables divided by the number of words).
    \item The Coleman–Liau index uses letters instead of syllables. The formula takes into account the average number of letters per word and the average number of words per sentence.
     \begin{equation}
      \textit{R}_\textit{C} = 0.0588 \cdot L - 0.296 \cdot S - 15.8,
    \end{equation}
    where \textit{L} -- average number of letters per 100 words,
    \textit{S} -- average number of sentences per 100 words.
    \item The ARI formula takes into account the number of letters. In the past, this allowed the use of this index to measure the complexity of texts in real time in electric typewriters.
     \begin{equation}
      \textit{R}_\textit{A} = 4.71\cdot \frac{characters}{words} + 0.5\cdot \frac{words}{sentences} - 21.43,
    \end{equation}
    where \textit{characters} -- number of letters and numbers,
    \textit{words} -- number of words,
    \textit{sentences} -- number of sentences.
    \item The main idea of the SMOG grade is that the complexity of the text is most affected by complex words. Complex words are words with many syllables (more than 3). The more syllables the more complicated the word.
    \begin{equation}
      \textit{R}_\textit{S} = 1.043\cdot \sqrt{polysyllable\cdot \frac{30}{sentences}}+3.1291,
    \end{equation}
    where \textit{polysyllable} -- number of polysyllable words,
    \textit{sentences} -- number of sentences.
    \item The Dale-Chall formula uses a count of ''hard'' words. These ''hard'' words are words that do not appear on a specially designed list of common words familiar to most 4th-grade students.
    \begin{equation}
      \textit{R}_\textit{D} = 0.1579\cdot \frac{difficult}{words}\cdot 100 + 0.0496\cdot \frac{words}{sentences},
    \end{equation}
    where \textit{difficult} -- number of difficult words,
    \textit{words} -- number of words,
    \textit{sentences} -- number of sentences.
\end{itemize}

In addition to the above, there are many other readability tests that are also actively used, e.g. the Fry Graph readability formula, the Spache index, the Linsear Write formula and others. The values obtained from readability tests are called readability indices. The Readability Index characterizes the difficulty of perceiving a text or the expected level of education that is required to understand it.

The readability formulas listed above are metrics for English texts. At the same time, the quantitative characteristics of other languages can differ significantly. For instance, Russian sentences are on average shorter than English, and words are longer. Therefore, the readability formulas need to be processed for use in other languages. Up to now, several studies have suggested the adaptation of readability tests for Russian. For example, I. Oborneva \cite{Oborneva} proposed the coefficients for the Flesch–Kincaid formula for Russian texts. The project \cite{readability} offers the adaptation of several readability formulas. M. Solnyshkina et al. presented a new approach to reading difficulty prediction in Russian texts \cite{Solnyshkina,Solovyev}. 

Readability is however only one aspect of age-based classification. Scholars have proposed more complex techniques for text complexity estimation using features of different nature. Thus, Yu. Tomina \cite{Tomina} considered the lexical and syntactic features of the text complexity level. A. Laposhina et al. \cite{Laposhina} evaluated a wide range of different types of features, such as readability, semantic, lexical, grammatical and others. M. Shafaei et al. \cite{Shafaei} estimated age suitability rating of movie dialogs using genre and sentiment features. L. Flekova et al. \cite{Flekova} proposed an approach to describing the story complexity for literary text. Y. Bertills \cite{Bertills} wrote about the features of literary characters and named entities in books for children. Finally, in our previous research, we evaluated the informativeness of some quantitative and categorical features for age-based text classification \cite{Glazkova}. The modern methodology for text difficulty estimation is based in most cases on machine learning approaches. Thus, R. Balyan et al. \cite{Balyan} showed that applying machine learning methods increased accuracy by more than 10\% as compared to classic readability metrics (e.g., Flesch–Kincaid formula). To date, a number of studies confirmed the effectiveness of various machine learning techniques for text difficulty estimation, such as support vector machine (SVM) \cite{Schwarm,Sung}, random forest \cite{Mukherjee}, and neural networks \cite{Azpiazu,Cuzzocrea,Schicchi}.

Another aspect of assessing the age category of text readers is the safety of the information it contains. Currently, in many countries, publishers are required to label books (including fiction) and other informational sources \cite{ars3,ars5,ars4,ars2,ars1} according to their age rating. For these purposes, there are special laws that rank information in terms of the potential harm it can bring. So, in Russia there is a Russian Age Rating System (RARS).

The RARS was introduced in 2012 when the Federal law of Russian Federation no. 436-FZ of 2010-12-23 "On Protection of Children from Information Harmful to Their Health and Development" was passed \cite{law}. The law prohibits the distribution of "harmful" information that depicts violence, unlawful activities, substance abuse, or self-harm. The RARS includes 5 categories, such as for children under the age of six (0+), for children over the age of six (6+), for children over the age of twelve (12+), for children over the age of sixteen (16+), and prohibited for children (18+). As a rule, an age rating is assigned to a book by editors or experts. As far as we know, there are currently no published research of how age rating correlates with other attributes of text, such as readability.

The reviewed studies and sources clearly indicate that age-based classification of fiction texts includes several aspects. First, the research topic is related to works on text difficulty evaluation. Text difficulty is characterized by different features, these are lexical, semantic, grammatical and other types. However, the measure of the difficulty of the text does not guarantee that this text is targeted to a particular age audience. It is required to evaluate the effectiveness of the existing text difficulty features for age-based classification. In addition, it would be interesting to evaluate the role of publishing attributes (for example, age rating labels) as classification features. Finally, the studies presented thus far provide evidence that machine learning approaches show the highest results in the task of estimating texts by difficulty. Based on this, it is reasonable to evaluate the text features for age-based classification using machine learning methods.

\section{Feature Types}

According to the related works, we consider the following types of classification features.

\begin{enumerate}
    \item 
    \textbf{General features.} This type includes features that reflect the quantitative characteristics of the text:
    \begin{itemize}
        \item the average and median length of words (\textit{avg\underline{ }words\underline{ }len}, \textit{med\underline{ }words\underline{ }len});
        \item the average and median length of sentences (\textit{avg\underline{ }sent\underline{ }len}, \textit{med\underline{ }sent\underline{ }len}), e.g. average or median number of symbols in each sentence;
        \item the average number of syllables (\textit{avg\underline{ }count\underline{ }syl});
        \item the percentage of long words with more than 4 syllables (\textit{many\underline{ }syllables});
        \item the Type-Token Ration, TTR (\textit{ttr}) \cite{Templin}. The main idea of the metric is that if the text is more complex, the author uses a more varied vocabulary so there’s a larger number of unique words. So, the TTR's value is calculated as the number of unique words divided by the number of words. As a result, the higher the TTR, the higher the variety of words;
        \item the TTR for nouns (\textit{ttr\underline{ }n}), adjectives (\textit{ttr\underline{ }a}), and verbs (\textit{ttr\underline{ }v}). The values of TTR calculated separately for parts of speech;
        \item the NAV metric (\textit{nav}). The NAV metric is a TTR-based ratio of (TTR A + TTR N)/TTR V proposed in \cite{Solnyshkina}.
    \end{itemize}
    \item 
    \textbf{Readability features.} We used the readability formulas with the coefficients for the Russian language proposed by the project \cite{readability}. In this study, we evaluated five types of readability indices using the following metrics: the Flesch–Kincaid readability test (\textit{index\underline{ }fk}); the Coleman–Liau index  (\textit{index\underline{ }cl}); the ARI index (\textit{index\underline{ }ari}); the SMOG grade (\textit{index\underline{ }SMOG}); the Dale-Chall formula  (\textit{index\underline{ }dc}).
    \item
    \textbf{Lexical features.} In this category, we included features constructed by the evaluation of the text in accordance with frequency dictionaries. As frequency dictionaries, we used the lists of Russian frequency words presented in \cite{Lyashevskaya,Sharoff}:
    \begin{itemize}
        \item the percentage of words included in the list of 5000 most frequent Russian words (\textit{5000\underline{ }proc});
        \item the average frequency of the words included in the 5000 most frequent words (\textit{5000\underline{ }freq});
        \item the average frequency of words per 1 million occurrences (ipm, \textit{words\underline{ }fr});
        \item the average frequency of nouns, verbs, adjectives, adverbs and proper names per 1 million occurrences (\textit{s\underline{ }fr}, \textit{v\underline{ }fr}, \textit{adj\underline{ }fr}, \textit{adv\underline{ }fr}, \textit{prop\underline{ }fr});
        \item the average number of topic segments of the corpus\footnote{The frequency dictionary was created on the basis of the modern subcorpus of the Main Corpus and the Oral Corpus of the Russian National Corpus (1950-2007) \cite{RNC} with a total volume of 92 million tokens \cite{Lyashevskaya}.} where the word was encountered (out of 100 possible, \textit{words\underline{ }r});
        \item the average number of the corresponding topic segments for nouns, verbs, adjectives, adverbs and proper names (\textit{s\underline{ }r}, \textit{v\underline{ }r}, \textit{adj\underline{ }r}, \textit{adv\underline{ }r}, \textit{prop\underline{ }r});
        \item the average value of Juilland's usage coefficients (\textit{words\underline{ }d}). This Juilland's usage coefficient measures the dispersion of the word's subfrequencies over  $n$  equally-sized  subcategories of the corpus \cite{Juilland};
        \item the average value of Juilland's usage coefficients for nouns, verbs, adjectives, adverbs and proper names (\textit{s\underline{ }d}, \textit{v\underline{ }d}, \textit{adj\underline{ }d}, \textit{adv\underline{ }d}, \textit{prop\underline{ }d});
        \item the number of documents in the corpora in which a word occurs (averaged over the text, \textit{words\underline{ }doc});
        \item the average number of documents in the corpora in which a word occurs (for nouns, verbs, adjectives, adverbs and proper names, \textit{s\underline{ }fr}, \textit{v\underline{ }fr}, \textit{adj\underline{ }fr}, \textit{adv\underline{ }fr}, \textit{prop\underline{ }fr}).
    \end{itemize}
    \item
    \textbf{Grammatical features.} We evaluated the percentage of nouns, verbs, and adjectives (\textit{count\underline{ }n}, \textit{count\underline{ }v}, \textit{count\underline{ }a}).
    \item
    \textbf{Sentiment features.} These features obtained with Russian Sentiment Lexicon \cite{Loukashevich}. We separately evaluated the percentage of positive and negative words for each of the topic categories, these are opinion, feeling (private state), or fact (sentiment connotation) (\textit{neg\underline{ }opinion}, \textit{neg\underline{ }feeling}, \textit{neg\underline{ }fact}, \textit{pos\underline{ }opinion}, \textit{pos\underline{ }feeling}, \textit{pos\underline{ }fact}).
    \item
    \textbf{Publishing features.} Here we have included features based on publishing attributes, i.e. on the book characteristics assigned by an editor or publisher, such as age rating according to the RARS (\textit{age\underline{ }rating}) and TF-IDF scores for book abstracts.
\end{enumerate}

\section{Dataset}

For feature evaluation, we collected a dataset of fiction books published in Russian. Due to copyright restrictions, the full texts of the books are not publicly available. Therefore, we used a collection of previews presented in online libraries in the public domain. Typically, the preview is 5-10\% of the total book volume.

The corpus consists of 5592 texts of children's and adult book previews. We have divided the texts into two parts. The first part included 4492 texts. It was used to train the models. The remaining 1000 texts were served as an independent text sample. The main characteristics of the data is presented in Table 1. Table 2 shows short text examples of adults and children's categories.

\begin{table}[h!]
\begin{center}
\caption{Characteristics of the corpus.}\label{tab1}
\begin{tabular}{|l|l|l|l|l|}
\hline
\multicolumn{1}{|l|}{\multirow{2}{*}{Characteristic}} & \multicolumn{2}{l|}{Training sample} & \multicolumn{2}{l|}{Test sample} \\ \cline{2-5} 
\multicolumn{1}{|l|}{} & \multicolumn{1}{|l|}{Children's} & \multicolumn{1}{|l|} {Adult} & \multicolumn{1}{|l|}{Children's} & \multicolumn{1}{|l|} {Adult} \\ 
\cline{1-5} 
Number of text & 2108 & 2384 & 500 & 500 \\
Avg number of symbols & 3134.38 & 3326.11 & 3048.69 & 3319.86 \\
Avg number of tokens & 488.55 & 499.52 & 479.3 & 498.16 \\
Avg number of sentences & 37.35 & 35.2 & 36.05 & 36.49 \\ \hline
\end{tabular}
\end{center}
\end{table}

\pagebreak
\begin{longtable}{|p{0.15\linewidth}|p{0.1\linewidth}|p{0.1\linewidth}|p{0.64\linewidth}|}
\caption{Example short fragments.}\label{tab2}\\
\hline
Category & Age Rating & Genre & Fragment \\ \hline
Adults & 16+ & Modern romance novels & A tall young man dressed in jeans, an inconspicuous jacket and a baseball cap pulled down with a visor over his eyes, approached the entrance of a seventeen-story apartment building and stood as if waiting for someone, and when a mother with a stroller appeared at the door, he quickly jumped inside - he did not know the code. I walked up to the fifth floor, putting on thin gloves on the go, looked around, and then deftly opened the door of one of the apartments. On the threshold he froze and listened for a while, but it was quiet. The man turned the baseball cap over the visor and began a leisurely survey of the apartment, opening the doors of the cupboards and looking into the drawers. The first thing he did was to open the sliding wardrobe door in the hallway, and oversized men's slippers, an empty box, and a bright purple scarf fell out onto the floor. The man grimaced and shoved everything back, muttering, "I thought so." In a large room that served as a bedroom, he lingered a little longer and grunted ironically at the sight of a luxurious couch with a carved back. \footnote{Fragment from the book "Men We Choose" by Evgeniya Perova (translated from Russian).}\\ \hline
Adults &  12+ &  Histori- cal adventures &  What a wonderful autumn it was in Southern Poland that year! Almost without rain and cold winds, tenderly warm, quiet, crimson-gold. Fabulous autumn - in such an autumn it is good, having climbed into the spurs of the Beskydy, from dawn to noon to wander along the slopes of hills overgrown with beech and hazel, and to your fill, drunk to breathe in the cool and crystal clear mountain air. And then, on the cozy terrace of a small mountain tavern, eat a good portion of hot, fiery-spicy bigos with pork legs, washed down with icy "okocim". And in the evening, having walked up to aching knees, kindle a fire on a platform open to all the winds above a shallow ravine, and, sitting on unbound logs, look at the stars that suddenly poured out in incredible numbers overhead. And, peering to the north, in the transparent thickening blue of the air, distinguish the lights of distant Krakow or Nova Huta, or maybe Bochnia or Wieliczka, who knows? \footnote{"Sold Poland" by Alexander Usovsky (translated from Russian).}\\ \hline
Children's & 6+ &  Child- ren's adventures & In a big, big city, where there are many, many houses, many, many cars and even more people, and the crows cannot be counted at all, there lived a ginger cat on a short street consisting of only two courtyards. His name was Ostrich. \footnote{"Greetings from cutlets" by Evgenia Malinkina (translated from Russian).}\\ \hline
Children's &  12+ & Child- ren's fantastic tales &  The hands of the clock were approaching half past seven, but the setting sun, reluctantly sliding behind the houses, continued to burn the city with rays, and the approaching twilight did not promise the long-awaited coolness.

Friday night was hot and stuffy, and the city roofs were so hot during the day that no sane cat would dare to run over them without burning their paws.

August was coming to an end, and the sun knew that it was the most important thing in the city, so from the very morning it climbed everywhere, trying to melt the asphalt on the streets, drying the grass on the lawns and sneaking into the apartments to flood them with heat and stuffiness. \footnote{"Vlad and the Secret Ghost" by Sasha Gotti (translated from Russian).}\\ \hline
\end{longtable}

Table 3 shows the most informative quantitative features with their means and standard deviation values. The informativeness is measured using the method of cumulative frequencies \cite{Aivazyan,Zagoruiko}. The main idea of this method consists in dividing the range of feature values for each class into $n$ intervals. The cumulative frequency of characteristic values is calculated for each interval. The informativeness indicator is calculated as the maximum absolute value of the difference in the accumulated frequencies for the corresponding intervals in the classes.

\begin{table}[]
\begin{center}
\caption{Top-10 of the most informative quantitative features (according to the method of cumulative frequencies).}\label{tab3}
\begin{tabular}{|l|l|l|l|l|}
\hline
\multicolumn{1}{|l|}{Feature} & \multicolumn{1}{|l|}{Mean (adult)} & \multicolumn{1}{|l|}{std (adult)} & \multicolumn{1}{|l|}{Mean (children's)} & \multicolumn{1}{|l|}{std (children's)} \\
\hline
\textit{avg\_sent\_len} & 105.65 & 54.51 & 88.69 & 30.64 \\
\textit{med\_sent\_len} & 97.9 & 59.06 & 79.69 & 34 \\
\textit{index\_dc} & 7.85 & 2.64 & 6.36 & 2.09 \\
\textit{adj\_doc} & 3484.28 & 1074.03 & 3669.48 & 1173.78 \\
\textit{index\_ari} & 9.29 & 3.53 & 7.4 & 2.93 \\
\textit{adj\_fr} & 135.3 & 53.56 & 146.96 & 60.39 \\
\textit{s\_doc} & 3705.73 & 838.45 & 3370.52 & 766.39 \\
\textit{index\_fk} & 9.21 & 3.66 & 7.36 & 2.95 \\
\textit{v\_doc} & 6255.7 & 2243.76 & 6481.53 & 1937.8 \\
\textit{adv\_fr} & 239.17 & 84.26 & 277.43 & 90.93 \\
\hline
\end{tabular}
\end{center}
\end{table}

Figure 1 presents the distribution of the age rating labels (age rating is a categorical feature) in the classes of the training data. It is interesting to note that some of the books from the children's class are labelled with the 18+ age category. Notable examples of this type of books are love stories for teens.

\begin{figure}[h]
\center{\includegraphics[scale=1]{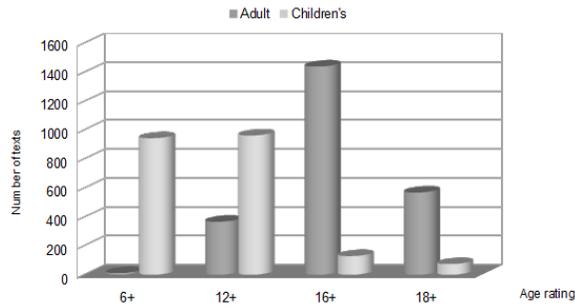}}
\caption{The distribution of age rating categories.}
\label{fig:image}
\end{figure}

\section{Experiments}

This section describes our feature evaluation experiments. We built two types of baseline models and sequentially enriched them with different types of features. Further, we compared the results obtained with our models with the results of CNN and RuBERT. Our dataset and models are available at \cite{github}.

\subsection{Baselines}

We built two classifiers for model evaluation. The first one was a Random Forest Classifier trained on bootstrap samples. The number of trees in the forest was equal to 100 and the Gini impurity was implemented to measure the quality of a split. The second model was a Linear Support Vector Classifier with the "l2" penalty and the squared hinge loss function. Both models were implemented using Scikit-learn \cite{Pedregosa} and Python 3.6.

\subsection{Preprocessing}

To preprocess our data, we used min-max normalization. Moreover, it is obvious that some of the features are correlated. For instance, most readability indices show a cross-correlation greater than 0.8. Another example of correlated feature pairs is average and median length of sentences or TTR values for all words and for particular parts of speech (Figure 2). To reduce the influence of feature correlation on the LSVC model, we applied linear dimensionality reduction using Singular Value Decomposition of the data with the minimum number of principal components such that 95\% of the variance is retained.

\begin{figure}[h]
\center{\includegraphics[scale=1]{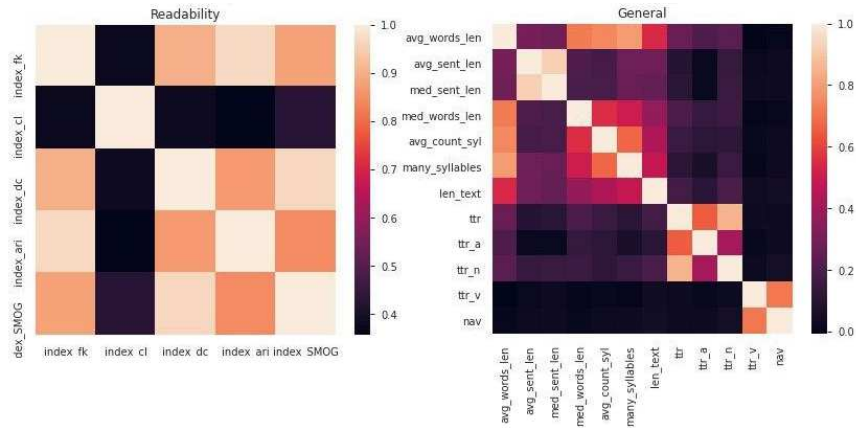}}
\caption{Correlation matrices for readability and general features.}
\label{fig:image}
\end{figure}

\subsection{Experiments and Results}

We used models trained of TF-IDF vectors as baselines. Further, we systematically evaluated each type of features.

To begin this process, we connected the TF-IDF vector of the text with the corresponding vector of features of a certain type. Book previews are rather long texts. Since the RuBERT model that participated in the comparison can only process a sequence of limited length, we used the same fragments of 256 tokens to train both neural networks and to construct TF-IDF vectors\footnote{The maximum sequence length for BERT is 512 tokens. However, due to the rather large volume of the corpus, we were also limited in computational resources.}. TF-IDF vectors were built over the top 2000 words ordered by term frequency across the corpus. At the same time, the values of additional features were calculated for the full preview texts.

To build TF-IDF vectors and CNN, the texts were pre-processed. The preprocessing included the following steps: special character removal, lowercase translation, lemmatization, stop-word removal. Text preprocessing was implemented using NLTK \cite{NLTK} and Pymorphy2 \cite{pymorphy2}.

\begin{table}[htbp]
\caption{Age-based classification results (\%).}\label{tab4}
\begin{tabular}{|l|l|l|l|l|}
\hline
Model & Accuracy & F1-score & Precision & Recall \\ \hline
\multicolumn{5}{|c|}{Baselines} \\ \hline
RF (TF-IDF) & 85.8 & 86.37 & 90 & 83.03 \\ 
LSVC (TF-IDF) & 83.7 & 84.01 & 85.6 & 82.47 \\ \hline
\multicolumn{5}{|c|}{Readability features} \\ \hline
RF baseline + readability & 86.5 & 86.91 & 89.6 & 84.37 \\ 
LSVC baseline + readability & 84.9 & 85.63 & 90 & 81.67 \\ 
RF (readability) & 60.6 & 61.6 & 63.2 & 60.08 \\ 
LSVC (readability) & 60.8 & 61.53 & 52.6 & 74.11 \\ \hline
\multicolumn{5}{|c|}{Sentiment features} \\ \hline
RF baseline + sentiment & 85.4 & 86.04 & 90 & 82.42 \\ 
LSVC baseline + sentiment & 83.8 & 84.09 & 85.6 & 82.63 \\ 
RF (sentiment) & 59.4 & 62.2 & 66.8 & 58.19 \\ 
LSVC (sentiment) & 68 & 67.01 & 65 & 69.15 \\ \hline
\multicolumn{5}{|c|}{Lexical features} \\ \hline
RF baseline + lexical & 83.2 & 84.23 & 89.9 & 79.23 \\ 
LSVC baseline + lexical & 84.2 & 84.45 & 85.8 & 83.14 \\ 
RF (lexical) & 63.1 & 64.62 & 67.4 & 62.06 \\ 
LSVC (lexical) & 61.8 & 62.91 & 64.8 & 61.13 \\ \hline
\multicolumn{5}{|c|}{Grammatical features} \\ \hline
RF baseline + grammatical & 85.5 & 86.26 & \textit{\textbf{91}} & 81.98 \\ 
LSVC baseline + grammatical & 83.7 & 84.03 & 85.6 & 82.47 \\ 
RF (grammatical) & 56.3 & 57.86 & 60.1 & 55.87 \\ 
LSVC (grammatical) & 59.6 & 62.1 & 66.2 & 58.48 \\ \hline
\multicolumn{5}{|c|}{General features} \\ \hline
RF baseline + general & 89.8 & 87.57 & 90.2 & 85.09 \\ 
LSVC baseline + general & 87.9 & 88.01 & 88.8 & 87.23 \\ 
RF (general) & 60.8 & 61.72 & 63.2 & 60.31 \\ 
LSVC (general) & 68.8 & 70.51 & 74.6 & 66.85 \\ \hline
\multicolumn{5}{|c|}{Publishing attributes} \\ \hline
RF baseline + abstracts & 87.4 & 87.77 & 90.41 & 85.28 \\ 
LSVC baseline + abstracts & 85.7 & 85.89 & 87.01 & 84.8 \\ 
RF baseline + age rating & 90.1 & 90.06 & 91.21 & 88.93 \\ 
LSVC baseline + age rating & 89.9 & 88.94 & 88.42 & 89.46 \\ 
RF baseline + publ. attr. & 90.4 & 90.94 & 96.4 & 86.07 \\
LSVC baseline + publ. attr. & 93 & 92.9 & 91.6 & 94.24 \\ \hline
\multicolumn{5}{|c|}{All features} \\ \hline
RF baseline + all features & 94.9 & 94.83 & 93.6 & 96.1\\ 
LSVC baseline + all features & \textbf{95.8} & \textbf{95.77} & \textbf{95} & \textbf{96.54} \\ 
RF (all features) & 94.7 & 94.67 & 94.2 & 95.15 \\
LSVC (all features) & 94.2 & 94.09 & 92.4 & 95.85 \\ 
RF baseline+all features–publ.attr. & 86.1 & 86.41 & 88.4 & 84.51 \\
LSVC baseline+all features–publ.attr. & 87.3 & 87.54 & 89.2 & 85.93 \\ \hline
RuBERT & \textit{\textbf{90.5}} & \textit{\textbf{90.16}} & 84.28 & \textit{\textbf{93.55}} \\
FNN (RuBERT embs + age rating) & 94.8 & 94.78 & 94.4 & 95.31 \\ \hline
CNN & 82.1 & 80.2 & 89.6 & 72.59 \\ \hline
\end{tabular}
\end{table}

Table 4 shows the results obtained for each type of features (e.g. RF baseline + readability, LSVC baseline + readability) and the results of the models trained only on additional features without TF-IDF vectors (e.g. RF (readability), LSVC (readability)). For publishing attributes, we evaluated two separate types of models. The first type used book abstracts as supplementary information. In other words, we added the texts of abstract to the book preview and built new TF-IDF vectors. The second type used baseline TF-IDF vectors with an additional feature of age rating. Finally, we evaluated three types of combined models, such as using all considered features, only all additional features, and all features with the exception of editorial attributes. The results obtained were compared with the results of three neural models:
\begin{itemize}
    \item RuBERT \cite{Kuratov}, based on BERT architecture \cite{Devlin}. BERT showed state-of-the-art results on a wide range of NLP tasks. RuBERT was trained on the Russian part of Wikipedia and news data. The model was implemented using PyTorch \cite{Paszke} and Transformers \cite{Wolf} libraries, it was trained for 3 epoches;
    \item FNN trained on fine-tuned RuBERT \cite{Kuratov} text embeddings obtained with PyTorch library \cite{Paszke}. Text embeddings were calculated by averaging the token vectors of the last hidden state. Age rating was presented as a one-hot numeric array which is the most widely used coding scheme \cite{coding}. FNN consisted of three layers including an input layer, a 1024 hidden layer with hyperbolic tangent activation function, and an output layer with softmax activation function. We also used Adam as an optimizer and binary cross-entropy loss. The FNN model was implemented using Keras \cite{Keras} library;
    \item CNN trained on Word2Vec embeddings \cite{Kutuzov}. CNN consisted of four building units including three convolutional units (CU) and a fully connected unit. Each CU contained the following sequence of layers $C-BN-C-BN-P$ where $C$ is a convolutional layer (CL), $BN$ is a batch normalization layer, $P$ is a pooling layer. After every CL the LeakyRelu activation function was applied. At the first CU we used 512 filters $7\times7$. At the second CL we applied 1024 filters $5\times5$. As a pooling strategy at each layer we used max polling with a kernel $2\times2$. The fully connected layer consisted of the following sequence of layers $FL_1-BN-FN_2$ where $FL_1$ is a hidden layer with 32 neurons, $FL_2$ is an output layer. We applied ReLU as an activation function and used stochastic gradient descent with Nesterov momentum and learning rate equal to $5\times10^{-2}$ as optimization parameters. The model was implemented with PyTorch \cite{Paszke}.
\end{itemize}

The results show that additional features in most cases improve the quality of baselines. According to F1-score, this concerns readability features, general features and publishing attributes. We assume that the advantage of these features is that they describe the text at the document level and allow the model to evaluate the whole text, and not just a fragment. It also can be seen that the using of abstracts and the age rating feature significantly improves the quality of the classification. The best results was obtained by the LSVC model using all considered features (95.8\% of accuracy, 95.77\% of F1-score, 95\% of precision, and 96.54\% of recall). These values are shown in bold in Table 3. Among the models that did not use publishing attributes, the best results were shown by RuBERT (90.5\% of accuracy, 90.16\% of F1-score, and 93.55\% of recall) and the LSVC baseline with grammatical features (91\% of precision).

\section{Conclusion}

The purpose of the current study was to evaluate different types of features for the task of age-based text classification. The results of this investigation show that features used in text difficulty evaluation can improve the quality of age-based classification. In addition, in this study, we considered publishing attributes (such as book abstracts and age ratings) as classification features. The results showed that the use of these attributes in digital libraries and recommendation systems could significantly improve the quality of machine learning approaches. Our further research will focus on studying other types of features, such as named entity analysis or plot and character features.

\end{document}